# Exploring the stimulative effect on following drivers in a consecutive lane-change using microscopic vehicle trajectory data

Ruifeng Gu [a], Ye Li [a, *], Xuekai Cen [a]

a. School of Traffic and Transportation Engineering, Central South University, Changsha, Hunan 410075, P. R. China

**Abstract:**

Improper lane-changing behaviors may result in breakdown of traffic flow and the occurrence of various types of collisions. This study investigates lane-changing behaviors of multiple vehicles and the stimulative effect on following drivers in a consecutive lane-changing scenario. The microscopic trajectory data from the HighD dataset are used for driving behavior analysis. Two discretionary lane-changing vehicle groups constitute a consecutive lane-changing scenario, and not only distance- and speed-related factors but also driving behaviors are taken into account to examine the impacts on the utility of following lane-changing vehicles. A random parameters logit model is developed to capture the driver's psychological heterogeneity in the consecutive lane-changing situation. Furthermore, a lane-changing utility prediction model is established based on three supervised learning algorithms to detect the improper lane-changing decision. Results indicate that (1) the consecutive lane-changing behaviors have a significant negative effect on the following lane-changing vehicles after lane-change; (2) the stimulative effect exists in a consecutive lane-change situation and its influence is heterogeneous due to different psychological activities of drivers; and (3) the utility prediction model can be used to detect an improper lane-changing decision.

# 1. INTRODUCTION

Improper lane-changing behaviors may result in the breakdown of traffic flow and reduced capacity [1] and the increased risk of traffic crashes [2] . The number of motor vehicle crashes due to sideswipe (related to improper lane-change) is approximately 863,100 in the United States in 2018 [3], among which a large proportion of crashes are caused by decision errors of drivers. Therefore, a number of studies have been conducted to investigate lane-change maneuvers from the perspective of driving behaviors [4,5] .

The lane-changing behaviors can be divided into discretionary lane-change and mandatory lane-change [6] . Drivers in a discretionary lane-change purse benefits such as faster speeds and more comfortable and safer driving environment, while mandatory lane-changing drivers change their lanes to drive along a preset path. A well-known theory for discretionary lane-change research is the utility-based lane choice. The utility-based theory has been extensively applied for exploring lane-changing behaviors in previous studies [7,8] . In the theory, drivers will compare utilities of different lanes and choose the lane with the highest utility and then perform lane-changes.

An important question, however, has not yet been answered, i.e., does the discretionary lane-change drivers always obtain a higher lane utility after changing their lanes? This issue is worthy of being investigated since it is related to the fundamental driving behavior about whether a lane-changing decision of a driver is correct and proper. If improper lane-changing behaviors could be avoided, traffic efficiency and safety may be improved accordingly. Previously, the issue was difficult to be examined due to the lack of empirical data. Prior studies mostly conducted off-road experiments (such as driving simulators) to explore driver's psychological and behavioral reaction to physical environment [9,10,11] , but it was difficult to account for actual conditions (such as lighting changes, shadows, vibrations and occlusions) that are encountered in real-world situations.

Recently, with the advancement of traffic detection and video surveillance technologies, naturalistic trajectory data become available [11,12,13,14,15,16,17] . The new data collection methods have an advantage of allowing a direct observation of driver's behavior and provide larger-scale datasets compared to driving simulators [18] . More behavioral and psychological factors in traffic scenarios may be also examined. Therefore, this study aims to investigate the aforementioned question by focusing on a special consecutive lane-changing situation using the real trajectory data. Considering two drivers perform discretionary lane-change behaviors consecutively, does the following lane-change driver obtain the higher utility by changing his/her lane, or just stimulated by the leading lane-change driver?

The primary objective of this study is to explore the stimulative effect in the consecutive lane-changing situation as well as the influencing factors of lane-changing behaviors. A microscopic trajectory dataset, the HighD dataset [19], is employed to extract real vehicle trajectories of consecutive lane-changes. The utilities of lane-changes are calculated and compared based on microscopic trajectory data. A random parameters logit model is developed to capture the driver's psychological heterogeneity in the consecutive lane-changing situation. The lane-changing utility prediction model is further established based on three supervised learning algorithms to detect the improper lane-changing decision.

The remainder of this study is organized as follows. The literature review is presented in Section 2. Section 3 describes the dataset and methodology, followed by the results and discussion of model estimation in Section 4. Section 5 concludes the major findings and provides future research directions.

## 2. LITERATURE REVIEW
### 2.1 Drivers' psychology and behavior in lane-change

There are two approaches commonly used to explain drivers' psychological and behavioral reactions in different lane-changing situations, i.e., off-road experiments and field studies. Off-road experiments (such as driving simulators and motion tracking devices) have been a mature approach to test how driving behaviors will be influenced by different physical environment [20,21,22]. For example, Yuan et al. [23] used a driving simulator to conduct an experiment with 54 licensed-drivers to study mandatory lane-changes in a weaving segment of a managed lane, and the authors found that middle-aged drivers have significantly shorter lane-changing duration than the elder ones. Huo et al. [24] used an eye tracker device to collect the eye movement markers and reaction time data of 35 drivers in Dalian, China. The result indicates that angry-driving-style participants' relative lane-changing decision time is shorter than that of other drivers.

Field studies about lane-changes are limited because real vehicle tests on road is costly and may result in safety issues. The nature of field studies is to collect real data of driving behaviors. GPS data owns the features of unobtrusive data collection, large-scale of the sample and real-time continuous dataset compared to traditional field studies methods [25]. Research on identifying driving style and behavior based on GPS data has received attention in recent years [26,27,28,29]. Recently, an alternative method makes trajectory data more accurate and informative, which utilizes microscopic vehicle trajectory data extracted from traffic videos. To the best of our knowledge, few studies have used microscopic trajectory data to investigate the relationship between the specific traffic behavior and psychology in traffic scenario so far [30,31].

### 2.2 Factors affecting lane-changing decision

Factors influencing drivers' lane-changing decision can be divided into two categories: real-word situations and driver's characteristics. Real-word situational factors (including traffic density, weather, light conditions, vehicle types, etc.) affect lane-changing decisions via drivers' gap acceptance [32]. Lee et al. [33] investigated the relative velocity between lanes and relative lead gap, and found that the relative lead gap has a positive effect on lane-changing probability, indicating that drivers tend to change into lanes with faster moving speeds or larger lead gaps. Li et al. [34] analyzed distance-related and speed-related factors of discretionary lane-changing vehicle groups during dynamic lane-changing decision-making process, and found that the lane-changing decision is a dynamic process where drivers gradually select more critical factors and pay more attention to them.

In the term of driver's characteristics, several studies were conducted regarding the relationship between sociodemographic feature and driving behavior [35,36,37]. Bener &

Crundall [36] found that young males tend to be more involved in risky driving behaviors. Additionally, driving cognitive demand and driver's age were found to significantly impact the lane-changing decision. The research revealed that older adults adopt a more conservative driving style during lane-changes and cognitive workload reduces the frequency of lane-changes for all age groups [37]. The study of Peplinska et al. [35] revealed that a variety of sociodemographic variables (such as age, gender, and the duration of having a driving-license) have the statistically significant correlation with dangerous driver syndrome.

**2.3 Lane-changing decision model**

The most well-known lane-changing decision model is based on the utility theory, in which drivers are assumed to always compare lane utilities and choose the highest one. Two factors are usually taken into consideration in lane utility, i.e., relative speed and gap acceptance [6] . Along this theory, numerous models have been proposed in past decades. For example, [38] adopted utility theory to model the decision process of lane-changes incorporating driver heterogeneity and state dependence. Toledo et al. [39] proposed an integrated lane-changing model where mandatory and discretionary conditions were joined together in a single utility model. Sun and Elefteriadou [40] conducted a focus group study to identify and understand drivers' concerns and responses under various lane-changing situations with the utility-based lane-changing decision model.

Recently, with the advancement of data collection techniques, microscopic trajectory data have been publicly available and more studies have been conducted based on the empirical data [41,42,43] . Xie et al. [44] proposed a data-driven lane-changing model based on deep learning algorithms. The results show that the most critical factor on lane-changing decision is the relative position of the preceding vehicle in the target lane. Gu et al. [45] raised the identification method of key point in the process of autonomous lane-change and established a vehicle lane-changing decision model base on random forest. These kinds of studies have been a hot topic since they have a sizeable impact on the design of advanced driver assistance systems, and consequently improving road safety [46] .

**3. DATA AND METHODOLOGY**

The overall research framework of data processing and methodology is shown in Fig. 1. The framework contains four phases in data processing and two phases in modeling analysis. Section 3.1 introduces the trajectory data of HighD dataset, and then the method of extracting lane-changing behaviors from trajectory data is described in Section 3.2. Lane-changing vehicle groups and consecutive lane-changing groups are proposed in Section 3.3. The purpose of Section 3.4 is to compare the utility of subject lane-changing vehicles obtained by lane-changing behaviors. Random parameters logit models (in Section 3.5) are used to explore the influencing factors of utility changes of following lane-changing vehicles and unobserved heterogeneity of drivers. In Section 3.6, supervised learning methods are applied to detect the improper lane-changing decision.

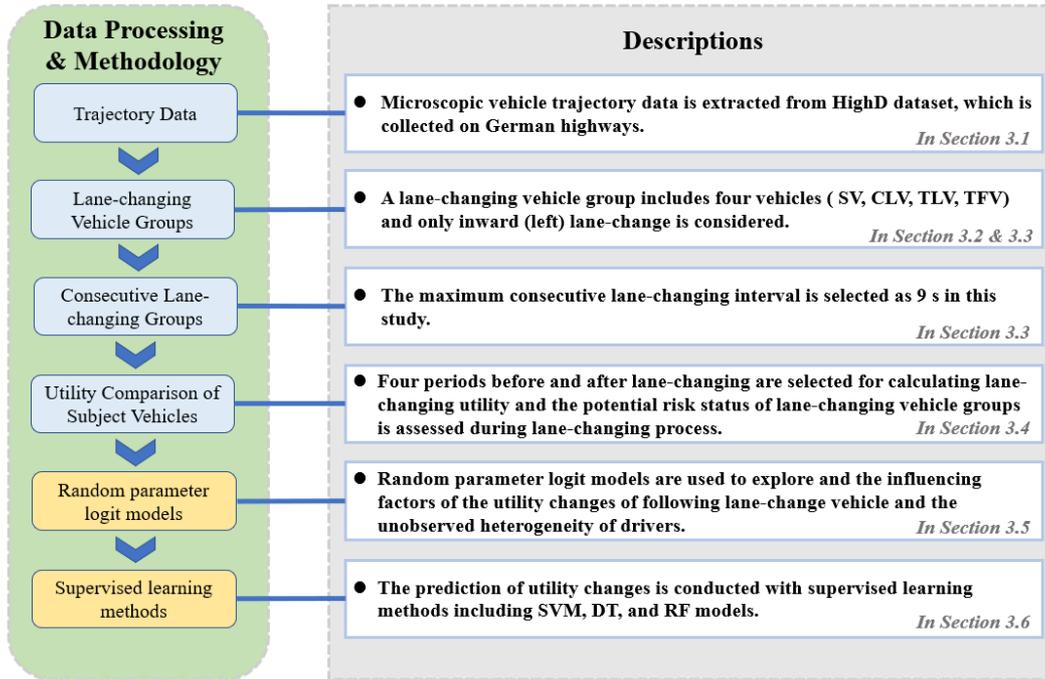

**Fig. 1. The framework of data processing and modeling analysis.**

## 3.1 Microscopic vehicle trajectory data

This study extracts microscopic vehicle trajectory data from the HighD dataset. The data are collected via camera-equipped drones from German highways. Using computer vision algorithms and manual annotation methods, the trajectory data are extracted from traffic videos, including frame, lane position, driving direction, lateral and longitudinal velocity, lateral and longitudinal acceleration, information of surrounding vehicles and others. The datapoints of vehicles are placed at the left rear corner of vehicles' bounding boxes (see Fig. 2).

The study sites of HighD include six different locations near Cologne, Germany, which are typical German highways with two or three lanes in each direction. The collection time of the HighD dataset is from 2017 to 2018. There are approximately 16.5 hours of trajectory data in the 60 sub-datasets. All lane-changing behaviors are considered as the discretionary lane-changes because ramps are not included in the analyzed sites. Besides, only inward lane-changing behaviors are included in this study, in order to ensure that lane-changing vehicles are pursuing the higher utility.

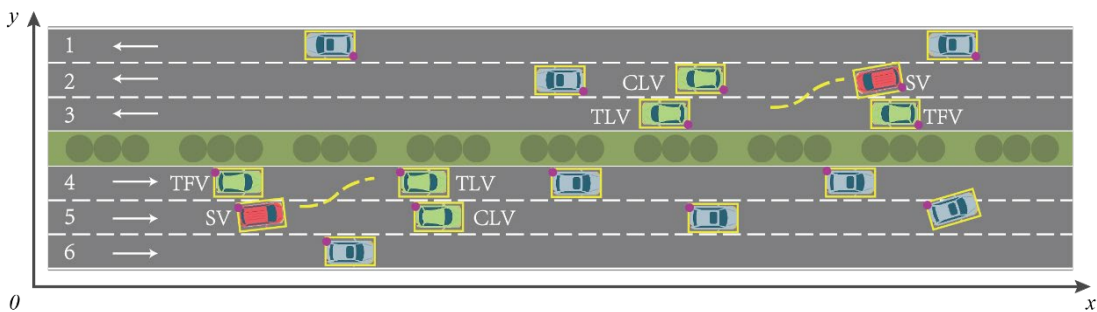

**(a)**

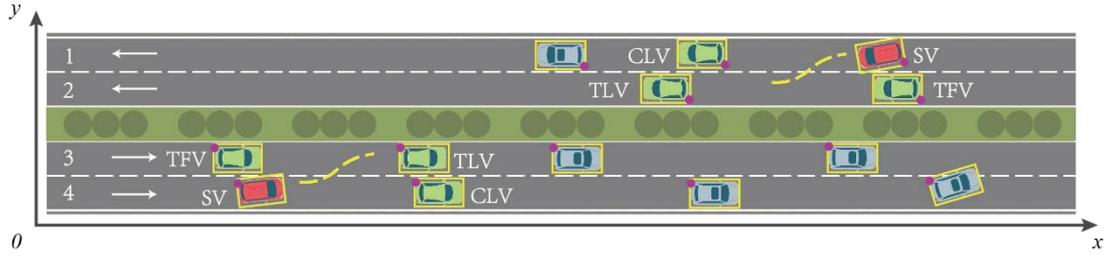

**(b)**

**Fig. 2. Illustration of the trajectory data collection site.**

### 3.2 Lane-changing operation

Before extracting the maneuvers of consecutive lane-changing behaviors, we need to pinpoint the time interval of each lane-changing operation, which can be calculated via the start time $t_s$ and end time $t_e$ of an operation process [14,34]. As shown in Fig. 3, The HighD dataset provides the position of the left rear point and the width $w$ of the vehicle, and the lateral position $D(line)$ of lane marking line. By comparing vehicle positions with the lane line positions, we can obtain the lane-changing start time $t_s$ and the end time $t_e$. In the present study, when the lane number of the lane-changing subject vehicle (SV) is changed in the dataset, we define $ID(t)$ as the lane ID used by the SV at time $t$. When $ID(t) \neq ID(t + \Delta t)$ ($\Delta t$ is the time interval between two consecutive data points), the lane ID changes and we define this moment $t$ as $t_{lc}$.

The second step is to calculate lateral position difference. We define $D(t)$ as the lateral position difference between the right rear point of SV and the left boundaries of the road. According to the dataset, the angles between the vehicle and the lane are very small, so the vehicles are assumed to keep parallel to the lane marking lines without considering the angles. Hence, we use $D(t)$ as the lateral position of the right boundaries of the lane-changing vehicle at time $t$.

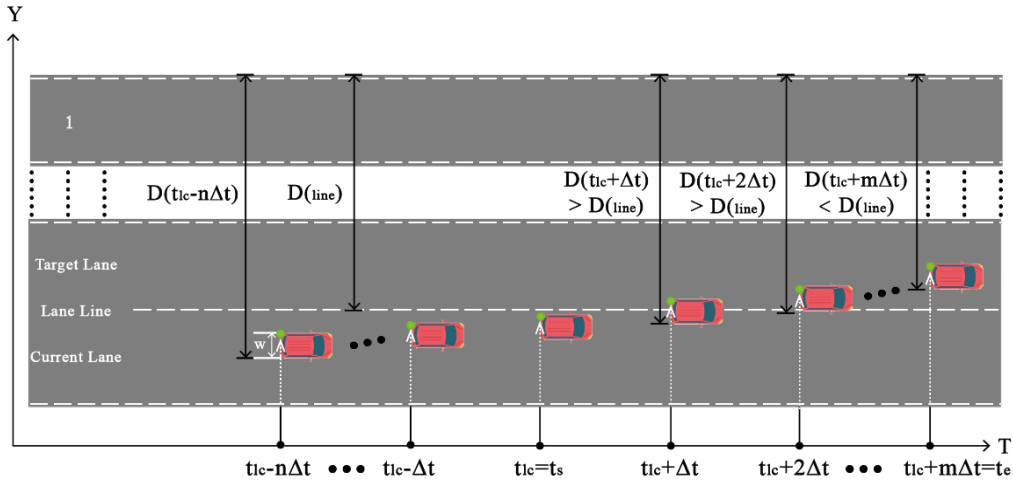

**Fig. 3. Illustration of the inward lane-changing operation.**

Further, we utilize the lateral position difference to calculate the lane-changing duration. When the vehicle executes inward (left) lane-change, the lane-changing start time $t_s$ is $t_{lc}$. After $t_{lc}$, $D(t)$ will still be larger than $D(line)$ for a while and then be smaller. The $t_e$ is the minimum $t$ when $D(t)$ is smaller than $D(line)$, which means the right boundary of the lane-changing vehicle has already pass the central line and driving into the target lane. It should be noted that the lane-changing duration we extracted only includes the period that the SV crosses the two lanes, which is the most critical period of a lane-changing process. The lane-changing start time $t_s$, the end time $t_e$ and the duration of lane-change $T_{LC}$ can be calculated as follows:

$$t_s = t_{lc} \tag{1}$$
$$t_e = min\ \{t|t > t_{lc}\ and\ D(t) < D(line)\} \tag{2}$$
$$T_{LC} = t_e - t_s \tag{3}$$

### 3.3 Consecutive lane-changing vehicle groups

In order to investigate whether lane-changing behaviors of preceding vehicle group may affect the following ones and compare the utility before and after lane-changes, consecutive lane-changing vehicle groups are extracted. Each lane-changing group involves four vehicles, including the leading vehicle in the target lane (TLV), the following vehicle in the target lane (TFV), and the leading vehicle in the current lane (CLV) as well as the lane-changing subject vehicle (SV). Vehicles of the group including the first lane-changing vehicle is collectively referred to as Vehicles1 (V1), and vehicles of the consecutive lane-changing vehicle group are collectively referred to as Vehicles2 (V2).

As shown in Fig. 4, after a preceding SV (SV1) executes a discretionary lane-change, the consecutive SV (SV2) driving in the same current lane also changes its lane to the same target lane. We defined this maneuver as a consecutive lane-change. In this maneuver, the time period between the start time of the first lane-changing SV ($t_{S1}$) and the start time of the consecutive lane-changing SV ($t_{S2}$) is denoted as the consecutive lane-changing interval. According to the previous studies, the maximum period of lane-changing impact lasts for 9 s [47,48], so the maximum consecutive lane-changing interval is selected as 9 s in this study. Then, data of two consecutive lane-changing vehicle groups are extracted to calculate the driving statuses and environment of SVs.

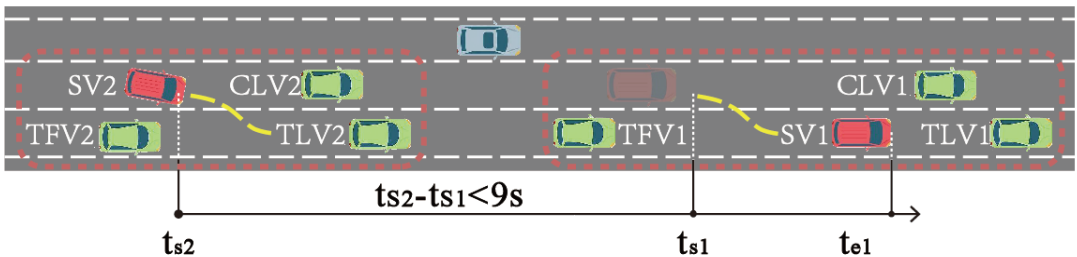

**Fig. 4. Illustration of consecutive lane-changing vehicle groups.**

### 3.4 Comparing utilities in consecutive lane-changing vehicle groups

The core question of this study is whether the following lane-changing SV2 gains a higher utility in a consecutive lane-changing situation or just influenced by the preceding lane-changing SV1. In order to quantify and compare utilities obtained by the following SV2 through lane-changing behaviors, we select vehicle spacing and speed as the safety and efficiency metrics respectively, which are two fundamental factors considered by lane-change drivers. By comparing driving states (including speed-related factors and distance-related factors) before and after lane-changes, utilities obtained by lane-changing behaviors can be quantified. Meanwhile, the conflict index, time-to-collision (TTC), is also applied to measure the potential risk during lane-changing processes for comprehensive evaluation of the stimulative effect on utilities.

Fig. 5 shows periods for comparison of lane-changing utilities. Four periods before and after the lane-changing behaviors are selected to quantify the utility changes after lane-changes. The $T_{LC}$ represents the lane-changing process of the SV2. The selection of the periods before lane-changes is based on the lane-changing start time of the following SV2 ($t_s$), and four periods forward are named $T_{1s}$, $T_{2s}$, $T_{3s}$, $T_{4s}$, respectively. Similarly, the selection of the periods after the lane-changes is based on the end time ($t_e$) of the lane-changing behavior of SV2, with the four periods denoted as $T_{1e}$, $T_{2e}$, $T_{3e}$, $T_{4e}$. The length of all eight periods before and after lane-changes is set as the half of the lane-changing duration (0.5$T_{LC}$).

For comparing utility obtained through lane-changing behaviors, each period before the lane-change ($T_{1s}$, $T_{2s}$, $T_{3s}$, $T_{4s}$) pairs a period after the lane-change ($T_{1e}$, $T_{2e}$, $T_{3e}$, $T_{4e}$) to constitute a sample set as to measure the utility change. For example, we compare the utilities of the following SV2 in the period of $T_{1s}$ and $T_{1e}$, and repeat the comparison for other periods. Table 1 shows sample sizes of each period's comparison set and the utility can be compared based on the data before and after lane-changes. Several utility measures based on distance and speed are examined to investigate impacts from two consecutive vehicle groups, which are displayed in Table 2. Note that, 472 qualified consecutive lane-changing scenario samples are extracted from the High-D dataset. But due to the limitation of the duration of HighD dataset, the data of some subject vehicles are not successfully collected for complete four-period length, resulting in different sample sizes of the period's comparison set. A total of 2,235 utility comparison set samples before and after lane-change of the consecutive lane-changing situations are extracted (see Table 1).

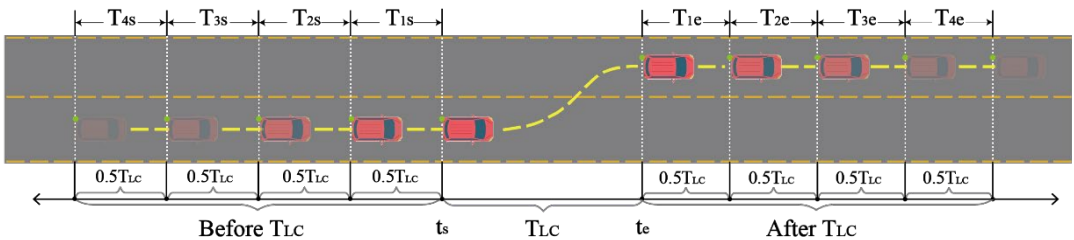

**Fig. 5. Illustration of periods for comparison of lane-changing utility of the following SV2.**

**Table 1: Sample sizes of the period's comparison sets.**

| Periods before lane-changes | Periods after lane-changes | | | |
|---|---|---|---|---|
| | $T_{1e}$ | $T_{2e}$ | $T_{3e}$ | $T_{4e}$ |
| $T_{1s}$ | 252 | 214 | 174 | 132 |
| $T_{2s}$ | 196 | 167 | 136 | 99 |
| $T_{3s}$ | 162 | 135 | 108 | 79 |
| $T_{4s}$ | 130 | 106 | 84 | 61 |

**Table 2: Utility measures.**

| Utility measures | Description |
|---|---|
| $\Delta Y_{difference}$ ((TLV2-SV2) – (TLV1-SV1)) | Clearance distance between TLV2 and SV2 minus clearance distance between TLV1 and SV1 after lane-changes (m) |
| $\Delta Y_{difference}$ ((SV2-TFV2) – (SV1-TFV1)) | Clearance distance between SV2 and TFV2 minus clearance distance between SV1 and TFV1 after lane-changes (m) |
| $\Delta V$ ($SV1_{after}$-$SV1_{before}$) | Vehicle speed difference of SV1 before and after lane-change (m/s) |
| $\Delta V$ ($SV2_{after}$-$SV2_{before}$) | Vehicle speed difference of SV2 before and after lane-change (m/s) |
| $\Delta V_{difference}$ (($SV2_{after}$-$SV2_{before}$) - ($SV1_{after}$-$SV1_{before}$)) | Difference between $\Delta V$ ($SV2_{after}$-$SV2_{before}$) and $\Delta V$ ($SV1_{after}$-$SV1_{before}$) (m/s) |

In order to compare potential risk during lane-changing process, the risk status of lane-changing vehicle groups is assessed based on the time-to-collision (TTC) index. The TTC index is an extensively utilized surrogate safety measure for risk assessments [49]. The TTC of the leading vehicles and following vehicles can be calculated as follows:

$$\text{TTC} = \begin{cases} \dfrac{X_{LV}-X_{FV}-\text{L}}{V_{FV}-V_{LV}}, & \text{if } V_{FV} > V_{LV} \\ \infty, & \text{if } V_{FV} \leq V_{FV} \end{cases} \quad (4)$$

where L is the length of the leading vehicle, $X_{LV}$ and $X_{FV}$ represent the position of the leading vehicle and its following vehicle respectively, and $V_{LV}$ and $V_{FV}$ are the velocity of the leading vehicle and its following vehicle respectively.

Whether two vehicles have potential risk is generally determined by comparing the TTC value with a threshold. The threshold values of TTC are ranging from 1.5 seconds to 4 seconds according to previous research [12], and this study selected 4 seconds as the threshold of TTC for capturing more risky samples. It is considered that a conflict exists when the TTC value is lower than the critical threshold. Therefore, the risk status between the leading and following vehicles can be shown as:

$$Risk_{vehicles} = \begin{cases} 1 \text{ (exsits risk)}, & \text{if } 0 < \text{TTC} < 4 \\ 0 \text{ (no risk)}, & \text{else} \end{cases} \quad (5)$$

For a lane-changing process, three vehicle pairs should be considered in a lane-changing group, i.e., CLV&SV, TLV&SV, as well as TFV&SV. Therefore, the potential risk of a lane-changing group is defined as:

$$Risk_{vehicle\ group}=\begin{cases}1\ (\text{exsits risk}), & \text{else} \\ 0\ (\text{no risk}), & \text{if } Risk_{CLV\&SV}=Risk_{TLV\&SV}=Risk_{TFV\&SV}=0\end{cases} \quad (6)$$

where $Risk_{CLV\&SV}$, $Risk_{TLV\&SV}$ and $Risk_{TFV\&SV}$ are the risk status of the three pairs of vehicles CLV&SV, TLV&SV, and TFV&SV respectively.

### 3.5 Random parameter logit model

This study aims to explore the association between the utility changes (before and after lane-changes) of the following SV2 and influencing factors. The unobserved heterogeneity of drivers is also considered in lane-changing processes. Considering the positive and negative changes in utilities, a binary random parameters logit model is applied in this study, which is estimated by the simulated maximum likelihood approach.

When the lane-changing behavior of the following SV2 has a positive utility change ($y = 1$), the probability is $p$ and the probability of negative utility ($y = 0$) is $1 - p$. Then:

$$logit(p_i) = \log\left(\frac{p_i}{1 - p_i}\right) = \boldsymbol{\beta}'\boldsymbol{X_i} + \varepsilon_i \quad (7)$$

where $\boldsymbol{X_i}$ is the vector of explanatory variables and $\boldsymbol{\beta}$ is a set of corresponding parameters. The term $\varepsilon_i$ is the random error term which is assumed to be identically and independently distributed.

Furthermore, the coefficients in the random parameters logit model are assumed to be randomly distributed with the formulation as follows:

$$\boldsymbol{\beta_i} = \boldsymbol{\beta} + \boldsymbol{\delta_i} \quad (8)$$

where $\boldsymbol{\delta_i}$ is the randomly distributed terms. The normal distribution is applied in this study for its good performance in statistical fit.

### 3.6 Supervised learning methods

The positive and negative utility change can also be formulated as a classification problem. Classifier is a supervised classification learning technique that takes the values of various features of an example and predicts the class label that the example belongs to [50]. Once trained, the classifier can determine how important the information contained by the features is to the label of the example, and this relationship is tested on the test dataset. In this study, support vector machine (SVM), decision tree (DT) and random forest (RF) are applied as the supervised learning models to predict whether the following SV2 can benefit from lane-changing behaviors in the maneuver of consecutive lane-changes.

Specially, the radial basis function (RBF) kernel is chosen in the SVM model because of the better robustness compared to linear or polynomial SVMs, and its capability of mapping both linear and nonlinear relationships by manipulating values of error weighting regularization factor C (cost) and kernel parameter $\gamma$ (gamma) [51]. The mathematical formula of RBF kernel is:

$$(x_i, x_j) = e^{-\gamma |x_i - x_j|^2} \quad (9)$$

where $x_i$ and $x_j$ represents two data points and $\gamma$ is a predefined kernel parameter.

RF uses bagging, a bootstrap aggregation technique, to build an ensemble of decision trees as a basic classifier [52]. To construct each tree, the recursive binary splitting method is adopted in which factors are selected to divide the data into different parts. Different criteria may be used to determine how to separate the data [53]. The Gini index criterion was used in this study as follow:

$$G = \sum_{K=1}^{K} P_K^m(1 - P_K^m) \tag{10}$$

where $P_K^m$ is the proportion of class $K$ and observations in node $m$.

The analysis is undertaken using the package of "sklearn" in Python software. The samples used in supervised learning models is split into training (80%) and test (20%) sets. The standardization and normalization of variables are also implied in the machine learning estimators. The dependent and independent variables used in the random parameters logit model and supervised learning methods are displayed in Table 3. The variables $V_{after}SV2$, $\Delta Y_{after}\text{TLV2-SV2}$ and $\Delta Y_{after}\text{SV2-TFV2}$ are not considered in modellings due to the high correlation with dependent variable $y$.

**Table 3: Variable description.**

| Variable | Description |
|---|---|
| **Dependent variable** | |
| y | A binary response variable, in which the positive speed-utility change of SV2 takes a value of 1 and otherwise takes a value of 0 |
| **Independent variables** | |
| $\Delta Y_{after}$ TLV1-SV1 | Vehicle clearance distance between TLV1 and SV1 after LC1 (m) |
| $\Delta Y_{after}$ SV1-TFV1 | Vehicle clearance distance between SV1 and TFV1 after LC1 (m) |
| $\Delta Y_{before}$ CLV1-SV1 | Vehicle clearance distance between CLV1 and SV1 before LC1 (m) |
| $\Delta Y_{before}$ CLV2-SV2 | Vehicle clearance distance between CLV2 and SV2 before LC2 (m) |
| $V_{before}$ SV1 | Speed of SV1 before LC1 (m/s) |
| $V_{before}$ SV2 | Speed of SV2 before LC2 (m/s) |
| $V_{after}$ SV1 | Speed of SV1 after LC1 (m/s) |
| ΔT | Time interval between the start time of the lane change of SV1 and the start time of the lane change of SV2 (s) |

*Note: LC1 and LC2 represent the lane-changing process of SV1 and SV2, respectively.*

## 4 RESULT AND DISCUSSION
### 4.1 Results of utility comparison

The results of utility comparison before and after lane-changes of SV2 are shown in Fig. 6. Fig. 6 (a) presents the proportion of achieving a higher utility (ΔV ($SV2_{after}$-$SV2_{before}$) > 0) of SV2 after lane-changes. For example, 0.377 in the upper left corner of Fig. 6 (a) means that 37.7 percent of SV2 achieve a higher utility in the after lane-changing period $T_{1e}$ compared with the before lane-changing period $T_{1s}$. Fig. 6 (b) shows the value of the average changes in utility of SV2. Table 4 presents the comparison of potential risk between consecutive lane-changing groups during lane-changing process.

Results in Fig. 6 may answer the aforementioned question whether does the discretionary lane-change drivers always obtain a higher lane utility after changing their lanes? In Fig. 6 (a), the largest proportion of achieving a higher utility of SV2 is 0.57, revealing that about half of SV2 fail to obtain a higher utility by the consecutive lane-changing behaviors. The tendency of the proportion, which is stable from $T_{1e}$ to $T_{4e}$, implies that the influence of the consecutive lane-changing behavior obviously lasts for a long period. As shown in Fig. 6 (b), it can be also found the average utility change of SV2 is not improved in most comparing period sets. The result demonstrates that the consecutive lane-change behavior may have a negative effect on driving utility after lane-changes.

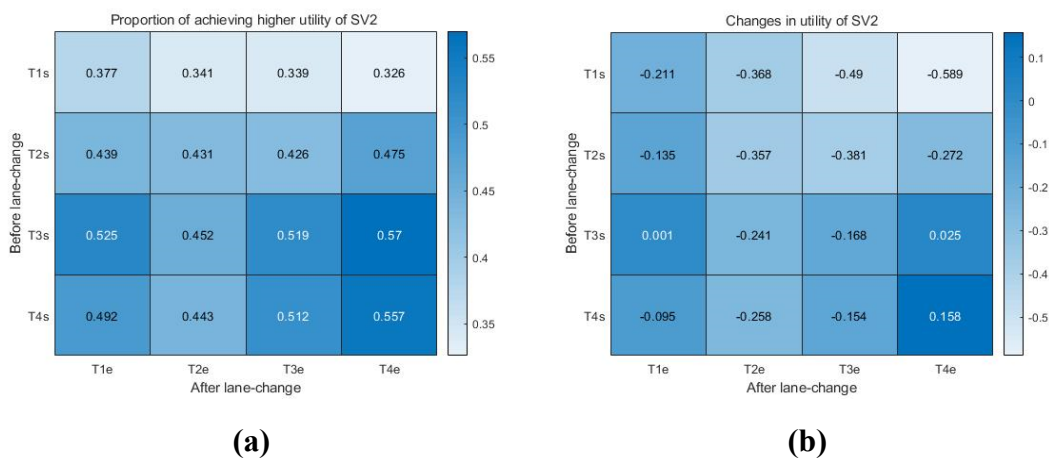

(a)          (b)

**Fig. 6. Utility comparison before and after lane-changes of SV2 in different periods.**

The utility change of SV2 and SV1 after lane-changes are also compared for analysis. The proportion of achieving relatively higher utility than SV1 ($\Delta V_{difference}$ (($SV2_{after}$-$SV2_{before}$) - ($SV1_{after}$-$SV1_{before}$))>0) of SV2 is shown in Fig. 7 (a), and the value of $\Delta V_{difference}$ (($SV2_{after}$-$SV2_{before}$) - ($SV1_{after}$-$SV1_{before}$)) is displayed in Fig. 7 (b). The trends of utility changes shown in Fig. 7 (a) and Fig. 7 (b) are similar to those in Fig. 6 (a) and Fig. 6 (b), respectively. Less than half of SV2 achieving higher utility than SV1 on average demonstrates that the consecutive lane-change has the negative effect.

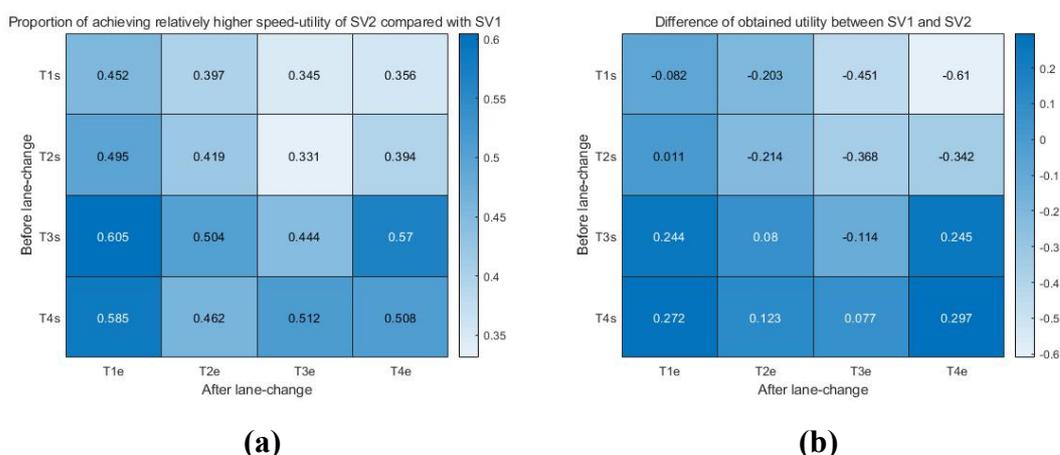

(a)          (b)

**Fig. 7. Utility comparison between SV2 and SV1 after lane-changes in different periods.**

Additionally, the clearance distance with other vehicles in the target lane is extracted to measure the safety-utility. Fig. 8 (a) and Fig. 8 (b) show the difference in safety-utility between SV2 and SV1 after lane-changes. $\Delta Y_{difference}$ ((SV2-TFV2) –(SV1-TFV1)) is the safety-utility measure of Fig. 8 (a) and the safety -utility measure of Fig. 8 (b) is $\Delta Y_{difference}$ ((TLV2-SV2) –(TLV1-SV1)).

As shown in Fig. 8 (a), in terms of safety-utility with TFV, SV2 is much lower than SV1 after lane-change, especially in $T_{1e}$ column. But the differences between SV2 and SV1 are reduced form $T_{1e}$ to $T_{3e}$. The result indicates that SV2 will have a more "crowded" driving environment with TFV than SV1 after lane-change, while the situation could become better with SV2's dynamic adjustment from $T_{1e}$ to $T_{3e}$. Besides, the difference in safety-utility with TLV between SV2 and SV1 is much larger than its with TFV (see Fig. 8 (b)), which indicates that the driving environment between SV2 and its TLV is worse than that with TFV. The previous research has proven that SV has a quite high risk of rear-end crash with TLV [14] and the consecutive lane-change behaviors may exacerbate this potential risk according to the result.

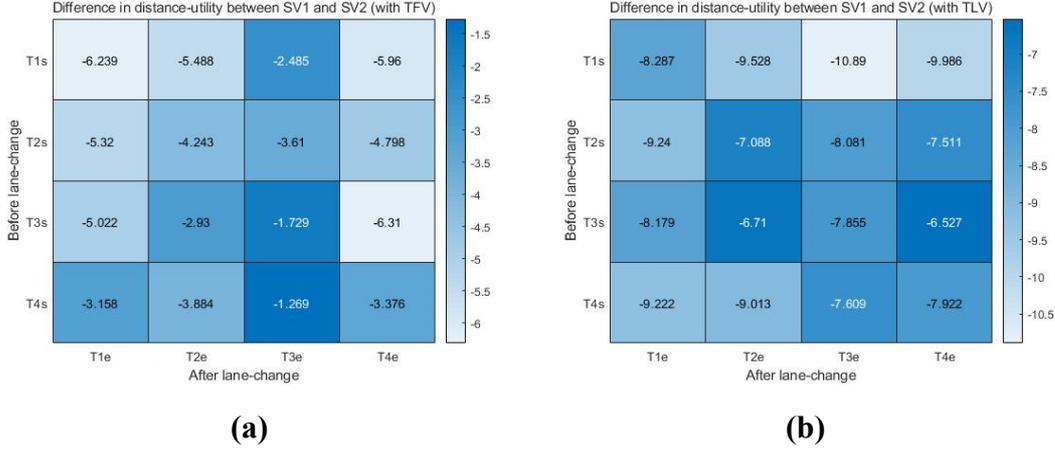

(a)          (b)

**Fig. 8. Safety comparison between SV2 and SV1 after lane-changes in different periods.**

When comes to the safety-utility during the lane-changing process, Table 4 illustrates that the preceding lane-changing vehicle groups (which include SV1) have higher potential risk status, with 13.136% of vehicle groups exist conflict, compared with the consecutive lane-changing vehicle groups (which include SV2). It indicates that the stimulative effect contributes to the safety of lane-change by providing a lane-changing example for the driver in advance. The interactions between SV and CLV are primarily responsible for resulting risk during lane-changing process.

**Table 4: Potential risk proportion of lane-changing vehicle groups during lane-changing process**

| Lane-changing vehicle | Surrounding vehicles | | | Vehicle group |
|---|---|---|---|---|
| | CLV | TLV | TFV | |
| SV1 | 12.712% | 0.212% | 0.212% | 13.136% |
| SV2 | 11.229% | 0.424% | 0.212% | 11.653% |

Overall, the utility analysis results show that discretionary lane-change drivers do not always obtain a higher lane utility after changing their lanes, and it is not a "good idea" for drivers to change their lanes in a consecutive lane-change situation.

**4.2 Results of the random parameters models**

Results of the random parameters logit models are presented in Table 5. Except for lane-changing intervals, variables considered in the models can be divided into two categories, including speed-related and distance-related factors. All the period's comparison sets are modeled and only three modelling results are significant, i.e., $T_{1S}$ vs. $T_{1e}$, $T_{3S}$ vs. $T_{1e}$, and $T_{4S}$ vs. $T_{1e}$. It can be found that the significant variables in the models are lane-changing interval time and distance-related factors, including the distance between SV1 and TFV1 after lane-changes ($\Delta Y_{after}$ SV1-TFV1) and the distance between SV2 and CLV2 before lane-changes ($\Delta Y_{before}$ CLV2-SV2).

Table 6 displays results of marginal effects. When $\Delta Y_{before}$ CLV2-SV2 increases by one unit, the probability of a higher utility acquired by SV2 will decrease by 0.5%, 0.7% and 0.7% for $T_{1S}$ vs. $T_{1e}$, $T_{3S}$ vs. $T_{1e}$, and $T_{4S}$ vs. $T_{1e}$, respectively. The larger $\Delta Y_{before}$ CLV2-SV2 means that the driving environment of current lane provides more space for SV2, and thus there would be fewer improvement of efficiency-utility (speed-utility) by lane-changes. Besides, in $T_{1S}$ vs. $T_{1e}$, when $\Delta Y_{after}$ SV1-TFV1 increases by one unit, the probability of getting a higher utility for SV2 will increase by 0.2%. However, the influence of $\Delta Y_{after}$ SV1-TFV1 to SV2 is not significant in period $T_{3s}$ and $T_{4s}$.

Additionally, we take the lane-changing interval of two vehicles (SV1 and SV2) as a variable into the models, and use the variable to measure the lane-changing behavior effect of SV1 on the lane-changing utility improvement of SV2. As shown in Table 5, the lane-changing interval $\Delta T$ is found to produce the random parameter in three models. When the lane-changing interval increases by one unit, there are 26.67%, 17.25% and 27.87% drivers having chances to gain higher utilities, respectively. At the same time, results of 73.33%, 82.75% and 72.13% drivers turn out to be opposite. It indicates that, with the influence of the lane-changing behaviors of SV1, most SV2's drivers (70-80%) still follow the utility-based rule to make lane-changing decisions, while a small part of SV2 drivers do not follow it. The findings demonstrate that different drivers have strong unobserved individual-specific heterogeneity when facing the lane-changing behaviors of other vehicles (i.e., the stimulation of lane-changing behavior) due to different psychological activities.

According to previous studies, the factors that lead to individual heterogeneity may include gender, age and driver-related behavior variables, etc. [54,55,56,57]. Therefore, we speculate the reason for this result is that the lane-changing behavior of SV1 may have a certain "stimulative effect", which will affect the judgment of consecutive vehicles' drivers on utilities. Some drivers may regard the lane-changing behaviors of SV1 as a guarantee for achieving a higher utility and make an irrational lane-changing decision.

Table 5: Results of the random parameters logit models.

| Variable | $T_{1S}$ vs. $T_{1e}$ | | $T_{3S}$ vs. $T_{1e}$ | | $T_{4S}$ vs. $T_{1e}$ | |
|---|---|---|---|---|---|---|
| | Parameter estimate | std. | Parameter estimate | std. | Parameter estimate | std. |
| $\Delta Y_{before}$ CLV2-SV2 | -0.022 | -0.022 | -0.029 | 0.006 | -0.027 | 0.008 |
| $\Delta Y_{after}$ SV1-TFV1 | 0.007 | 0.004 | - | - | - | - |
| Constant | 1.144 | 0.528 | 1.809 | 0.394 | 2.849 | 0.771 |
| **Random parameters (normally distributed)** | | | | | | |
| $\Delta T$ | -0.111 | 0.045 | -0.121 | 0.051 | -0.139 | 0.067 |
| Standard deviation of $\Delta T$ | 0.178 | 0.039 | 0.129 | 0.042 | 0.237 | 0.060 |
| **Model statistics** | | | | | | |
| LL(at convergence) $LL(\beta_r)$ | | -146.191 | | -97.453 | | -74.961 |
| LL(at constant) $LL(\beta_0)$ | | -166.967 | | -112.092 | | -90.094 |
| $\rho^2 = 1 - LL(\beta_r)/LL(\beta_0)$ | | 0.124 | | 0.131 | | 0.1689 |

Table 6: Results of marginal effects.

| Variable | $T_{1S}$ vs. $T_{1e}$ | $T_{3S}$ vs. $T_{1e}$ | $T_{4S}$ vs. $T_{1e}$ |
|---|---|---|---|
| $\Delta Y_{before}$ CLV2-SV2 | -0.005 | -0.007 | -0.007 |
| $\Delta Y_{after}$ SV1-TFV1 | 0.002 | - | - |
| $\Delta T$ | -0.026* | -0.030* | -0.035* |

*Note: * represents marginal effects for random parameters.*

## 4.3 Results of supervised learning methods

In this section, an exploratory research is performed on predicting whether drivers obtain a higher utility from the lane-changing behaviors. The lane-changing utility prediction model is established based on three supervised learning algorithms, which may be used to detect the improper lane-changing decision. In order to reduce the impact of the sample size and dynamic adaptive behaviors of drivers after lane-change on the prediction of utility, SVM, DT and RF models are established based on $T_{1s}\ vs.\ T_{1e}$ period's comparison set which contains 252 valid consecutive lane-changing cases. Table 7 shows the performance of the models to predict the utility obtained by SV2 in consecutive lane-changing situations. The prediction variable is set as a binary response variable, in which the value is taken as 1 when getting a higher utility, otherwise it is taken as 0.

According to Table 7, the accuracies of SVM and DT methods are less than 0.7, while the RF achieves the highest accuracy of 0.706. Confusion-Matrix and ROC curve of the prediction about the random forest are presented in Fig. 9. Note that, due to the limited sample size, there are only 201 and 51 samples in training and testing datasets, respectively. In the future, with more micro-trajectory data being applied into the RF model, the accuracy of prediction will be further improved.

**Table 7: Comparison results of supervised learning methods.**

|     | Train set (80%) | | | | Test set (20%) | | | |
| --- | --- | --- | --- | --- | --- | --- | --- | --- |
|     | Accuracy | Precision | Recall | F1 | Accuracy | Precision | Recall | F1 |
| **SVM** | 0.711 | 0.706 | 0.711 | 0.706 | 0.628 | 0.606 | 0.628 | 0.611 |
| **DT**  | 1.000 | 1.000 | 1.000 | 1.000 | 0.647 | 0.680 | 0.647 | 0.655 |
| **RF**  | 0.980 | 0.981 | 0.980 | 0.980 | 0.706 | 0.694 | 0.706 | 0.685 |

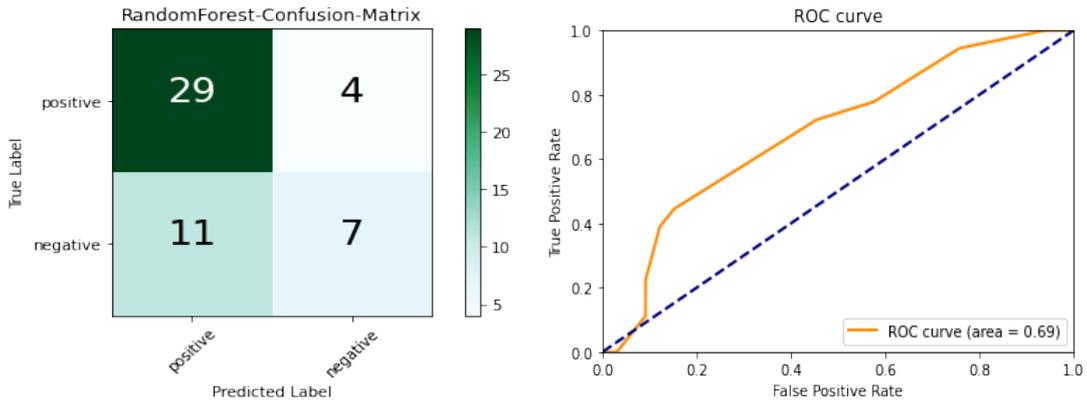

**Fig. 9. Prediction performance by random forest.**

The RF method can also estimate the importance of all input features in the classification. Fig. 10 and Table 8 presents the average importance with standard error of each variable. The importance of $\Delta Y_{before}$ CLV2-SV2 is obviously higher than other distance-related features, which is in accord with the estimated results in aforementioned random parameters models. The ΔT also plays a significant role in prediction.

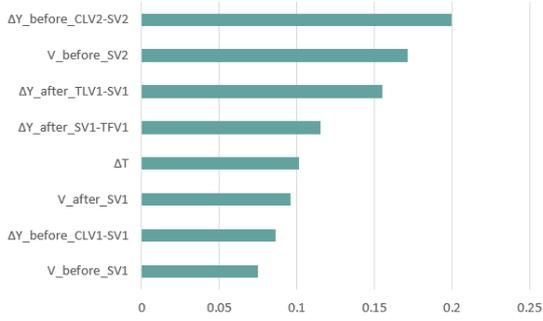

Fig. 10. Feature importance ranking.

Table 8: Feature importance ranking.

| Feature | Weight |
|---|---|
| $\Delta Y_{before}$ CLV2-SV2 | 0.199±0.228 |
| $\Delta V_{before}$ SV2 | 0.171±0.146 |
| $\Delta Y_{after}$ TLV1-SV1 | 0.155±0.152 |
| $\Delta Y_{after}$ SV1-TFV1 | 0.115±0.105 |
| ΔT | 0.102±0.113 |
| $\Delta V_{after}$ SV1 | 0.096±0.070 |
| $\Delta Y_{before}$ CLV1-SV1 | 0.086±0.102 |
| $\Delta V_{after}$ SV1 | 0.075±0.077 |

## 5. CONCLUSIONS

This study focused on lane-changing behaviors of multiple vehicles and the stimulative effect to following drivers in consecutive lane-changing situations, and analyzed the factors limiting the utility improvement of the following lane-changing vehicles (SV2). The random parameters logit model and supervised learning methods were employed using microscopic trajectory data from the HighD dataset. Two discretionary lane-changing vehicle groups constituted a consecutive lane-changing situation, and not only distance-related and speed-related factors but also behavior-related factors are taken into account to examine their impact on the utility of the following lane-changing vehicles. The major conclusions are as follows:

(1) Revealing the negative effect of the consecutive lane-changing behavior. No matter in efficiency-utility or safety-utility, the consecutive lane-changing behavior always has a negative effect after lane-changes according to the result of data analysis.

(2) Stimulative effect is found in the maneuver of consecutive lane-change. The significant random parameters of the lane-change interval variable ΔT quantitatively demonstrate that the psychological activities of drivers who witness the lane-changing behavior are heterogeneous. Some drivers may be influenced by stimulative effect from the leading lane-changing vehicles and make improper lane-changing decisions.

(3) Proposing the utility prediction method to detect the improper lane-changing decision. Three supervised learning methods (i.e., SVM, DT, RF) are tested and the accuracy of RF is the best (higher than 70%), proving the feasibility of utility prediction for the maneuver of consecutive lane-change.

The aforementioned findings reveal the existence of stimulative effect in the maneuver of consecutive lane-changes, and allow us to better understand the psychological activities of drivers in the process of lane-changing decision. Considering all the behaviors on the road, not the circumstances around the vehicle only, these conclusions can be utilized as references for intelligent vehicle infrastructure cooperative system to improve the stability of lane-changing decision and the accuracy of the driving intention prediction. Besides, the main factors limiting the improvement of the prediction model in accuracy is the insufficient samples. This problem will be solved when more micro-trajectory databases become available.

**ACKNOWLEDGMENTS**

This research was sponsored by : 1) the National Natural Science Foundation of China (No. 71901223); 2) Natural Science Foundation of Hunan Province (2021JJ40746); 3) the Postgraduate Research and Innovation Project of Central South University (No. 1053320216523)